\RequirePackage{amsmath}
\documentclass[a4paper,english]{article}
\usepackage[margin=1.5in,footskip=0.25in]{geometry}
\usepackage{amsfonts}
\usepackage{multirow}
\usepackage{subcaption}
\usepackage[T1]{fontenc}
\usepackage{graphicx}
\usepackage{authblk}
\usepackage{url}
 
\DeclareMathOperator{\E}{\mathbb{E}}
\DeclareMathOperator{\erf}{erf}
\DeclareMathOperator{\erfc}{erfc}
\DeclareMathOperator{\sign}{sign}
\begin{document}

\title{Flexible Tails for Normalising Flows, with Application to the Modelling of Financial Return Data}
\author[1]{Tennessee Hickling}
\author[1]{Dennis Prangle}
\affil[1]{
    University of Bristol, Bristol, United Kingdom
}
\affil[ ]{\textit{\{th17628,dennis.prangle\}@bristol.ac.uk}}
\maketitle

\begin{abstract}
We propose a transformation capable of altering the tail properties of a distribution, motivated by extreme value theory, which can be used as a layer in a normalizing flow to approximate multivariate heavy tailed distributions.
We apply this approach to model financial returns, capturing potentially extreme shocks that arise in such data.  
The trained models can be used directly to generate new synthetic sets of potentially extreme returns.
\end{abstract}

\section{Introduction} \label{sec:intro}
A normalising flow (NF) expresses a complex probability distribution as a parameterised transformation of a simpler base distribution.
A NF sample is
\begin{equation}
x = T(z; \theta),
\end{equation}
where $z$ is a sample from the base distribution, typically $\mathcal{N}(0,I)$.
A number of transformations have been proposed which produce flexible and tractable distributions.
Flexibility can be enhanced by composing multiple transformations. 
One popular application of NFs has been the generation of synthetic data, which can be performed following density estimation (fitting a transformation to observed data points) \cite{papamakarios_2021}.
NFs have achieved good performance in challenging high dimensional problems such as image \cite{kingma_2018} and audio generation \cite{prenger_2018}.
See \cite{kobyzev_2020} and \cite{papamakarios_2021} for reviews of NFs.

In \cite{jaini_2020}, Jaini et al prove that Lipschitz transformations leave tail behaviour of distributions unchanged (see Section \ref{sec:related_work} for a more precise statement).
Many normalizing flows use Lipschitz transformations of Gaussian base distributions, implying they produce distributions which retain Gaussian tails and cannot have heavier tails.
This is a crucial limitation when capturing heavy tails is critical, which arises naturally when studying exposure to extreme events.
For example, in many applications -- such as climate \cite{zscheischler_2018_climate_risk}, contagious disease \cite{cirillo_2020_disease_risk} and financial \cite{gilli_2006_financial_risk} --
the probability of very large events is often more relevant than the shape of the density around the mean. 
Research into ``market simulators'' (generation of synthetic financial time series), leveraged to expand training observations for data intensive approaches \cite{buehler_2018}, is an active area of research \cite{wiese_2020,wiese2021multiasset}.
In such cases, methods that produce realistic shocks (extreme events) are essential.

Based on the result of Jaini et al, many authors have proposed using heavy tailed base distributions.
We propose an alternative approach: using a Gaussian base distribution with a final non-Lipschitz transformation layer in $T$.
We propose a particular transformation motivated by extreme value theory.
Potential advantages of our approach are capturing asymmetry and complex dependence in tail weights.

In the remainder of this section we review related prior work.
Then Section \ref{sec:background} describes background material and Section \ref{ftt} states our proposed transformation.
Finally Section \ref{sec:experiments} illustrates its empirical performance in modelling high dimensional financial returns data.

\subsection{Related Work} \label{sec:related_work}

\paragraph{Theory}\mbox{}\\
Consider a univariate real valued random variable $Z$.
Foss et al \cite{foss_2011} define $Z$ as (right-)\textbf{heavy-tailed} when $\E[\exp(\lambda Z)] = \infty$ for all $\lambda > 0$
(examples include Student's $T$ and Pareto distributions), 
and \textbf{light-tailed} otherwise
(examples include the normal distribution). 
Using this definition, Theorem 2 of Jaini et al \cite{jaini_2020} proves that if $Z$ is light-tailed and $T$ is Lipschitz, then $T(Z)$ is also light-tailed.
In further results they derive the gradient required by $T$ to produce desired tail-properties, and generalise their results to the multivariate case.
Finally they also prove that several popular classes of NFs use Lipschitz transformations.

Other mathematical definitions of tail behaviour exist, including alternative usages of the term ``heavy tails''.
For instance \cite{liang_2022} prove similar results to those of Jaini et al using definitions based on theory of concentration functions.

To model heavy tailed distributions, Jaini et al note that there is a choice
``of either using source densities with the same heaviness as the target,
or deploying more expressive transformations than Lipschitz functions''.
They pursue the former approach and we investigate the latter.

\paragraph{Heavy Tailed Base Distributions}\mbox{}\\
Let $Z$ be a random vector with a normalising flow's base distribution, and $Z_i$ be its $i$th component.
Several NF papers propose a heavy tailed distribution for $Z$.
Firstly, in the setting of density estimation, \cite{jaini_2020} propose setting (independently)
$Z_i \sim t_\nu$,
a standard (location zero, scale one) Student's T distribution with $\nu$ degrees of freedom.
The $\nu$ parameter is learned jointly with the NF parameters.
Optimising their objective function requires evaluating the Student's $T$ log density
and its derivative with respect to $\nu$, which is straightforward.

An extension is to allow \textbf{tail anisotropy}
-- tail behaviour varying across dimensions --
by setting (independently) $Z_i \sim t_{\nu_i}$,
so the degrees of freedom can differ with $i$.
In \cite{laszkiewicz_2022}, two such approaches are proposed for density estimation.
The marginal Tail-Adaptive Flow (mTAF) approach first learns $\nu_i$ values,
and then fixes the distribution of $Z$ while training $\theta$ (which parameterises $T$).
A feature of mTAF is it allows Gaussian marginals by using Gaussian base distributions for the appropriate $Z_i$ components.
Generalised Tail-Adaptive Flows (gTAF) train the $\nu_i$s and $\theta$ jointly.
In the setting of variational inference, \cite{liang_2022} propose an Anisotropic Tail-Adaptive Flow (ATAF), which is similar to gTAF. 

\paragraph{Other Methods}\mbox{}\\
We also summarise two papers using alternatives to a Student's T base distribution,
in the setting of density estimation.
Firstly, \cite{amiri_2022} also consider a generalized Gaussian base distribution
(this has tails which can be heavier than Gaussian, but are lighter than Student's T)
and Gaussian mixture base distributions.
They argue that mixture distributions can in theory model any smooth density given enough components,
and also that they are more stable in optimisation than heavy tail base distributions.
Secondly, COMET flows \cite{mcDonald_2022} use a flow based copula distribution on the unit hypercube,
followed by marginal transformations based on the GPD distribution for the positive and negative tail.
The marginal transformation is fitted first, then fixed while the copula is learned.

\paragraph{Financial Simulation}\mbox{}\\
Previous data-driven simulation of financial data has applied generative adversarial networks in combination with a static transformation to lighten the tails of the data \cite{wiese_2020}.
This work has been extended to simulate multivariate financial returns using normalising flows \cite{wiese2021multiasset}, for which marginal transformations are learnt separately, similar to COMET flows.

\paragraph{Empirical Results}\mbox{}\\
All the related work listed above includes empirical examples,
providing evidence that allowing for extreme tails can improve performance in relevant applications.

\section{Background} \label{sec:background}

\subsection{Normalising Flows} \label{sec:flows} 
Consider vectors $z \in \mathbb{R}^d$ and $x = T(z) \in \mathbb{R}^d$.
Suppose $z$ is a sample from a base density $q_z(z)$.
Then the transformation $T$ defines a \textbf{normalising flow} density $q_x(x)$.
Suppose $T$ is a diffeomorphism (a bijection where $T$ and $T^{-1}$ are differentiable),
then the standard change of variables formula gives
\begin{equation} \label{eq:change_of_variables}
    q_x(x) = q_z(T^{-1}(x))|\det J_{T^{-1}}(x)|.
\end{equation}
Here $J_{T^{-1}}(x)$ denotes the Jacobian of the inverse transformation and $\det$ denotes determinant.

Typically a parametric transformation $T(z; \theta)$ is used,
and $q_z(z)$ is a fixed density, such as that of a $\mathcal{N}(0, I)$ distribution.
However some previous work uses a parametric base density $q_z(z; \theta)$.
(So $\theta$ denotes parameters defining both $T$ and $q_z$.)
For instance several methods from Section \ref{sec:related_work} use a Student's $T$ base distribution
with variable degrees of freedom.


Usually $T = T_K \circ \ldots \circ T_2 \circ T_1$, a composition of several simpler transformations.
Many such transformation have been proposed, with several desirable properties.
These include producing flexible transformations and allowing evaluation (and differentiation) of $T$, $T^{-1}$, and the Jacobian determinant.
Such properties permit tractable sampling, and density evaluation via \eqref{eq:change_of_variables}.
Typically, methods allow fast evaluation of the Jacobian determinant
and either $T$ (allowing fast sampling) or $T^{-1}$ (allowing fast density evaluation).
See \cite{kobyzev_2020,papamakarios_2021} for reviews of normalizing flows.

\subsection{Density Estimation}
Density Estimation aims to approximate a target density $p(x)$ from which we have a number of samples $\{x_i\}_{i=1}^N$.
As in Section \ref{sec:flows}, we assume $x \in \mathbb{R}^d$.

We can fit a normalising flow by minimising the objective
\begin{equation}
    \mathcal{J}(\theta) = -\sum_{i=1}^N \log q_x(x_i; \theta).
\end{equation}

This is a Monte Carlo approximation (up to a multiplicative constant)
of the Kullback-Leibler divergence $KL[p(x) || q_x(x; \theta)]$.
The objective gradient can be numerically evaluated using automatic differentiation.
Thus optimisation is possible by stochastic gradient methods.

A normalising flow is a convenient choice for $q_x(x; \theta)$, as it provides a flexible family of distributions and provides both sampling of $x \sim q_x$ and evaluation of $\log q_x(x; \theta)$.
Once trained, we can sample synthetic data by sampling the base distribution and applying our transformation.
As mentioned in Section \ref{sec:flows}, we can typically choose to make only one of density evaluation or sampling fast,
and in this paper we concentrate on the former.

\subsection{Extreme Value Theory} \label{sec:EVT}

Extreme value theory (EVT) is the branch of statistics studying extreme events \cite{coles_2001}.
A classic result is \textbf{Pickand's theorem} \cite{pickands_1975} (see \cite{papastathopoulos_2013} for a review).
Given a scalar real-valued random variable $X$, consider the scaled excess random variable $\frac{X-u}{h(u)} | X>u$,
where $u>0$ is a large threshold and $h(u)>0$ is an appropriate scaling function.
The theorem states that the scaled excess may only converge in distribution to a Generalized Pareto distribution (GPD)
(or a degenerate distribution).
A common EVT approach is to treat $h(u)$ as constant, and model tails of distributions as having GPD densities.

The GPD distribution involves a shape parameter or \textbf{tail index}, $\lambda \in \mathbb{R}$.
For $\lambda>0$, the GPD density is asymptotically (for large $x$) proportional to $x^{-1/\lambda-1}$,
while for $\lambda<0$ the upper tail has bounded support.
In terms of the definition of \cite{foss_2011} described in Section \ref{sec:related_work},
$\lambda>0$ produces heavy tails.
Given $X$, the tail index parameter of the GPD resulting from Pickand's theorem
is a measure of how heavy the tail of $X$ is.
A Gaussian distribution results in a tail index of zero,\footnote{
In terms of Pickands theorem, a Gaussian requires a non-constant scaling function, so it does not exactly have a GPD tail.}
and positive indices represent heavier tails.
Note that other related mathematical definitions of ``tail index'' exist.
See \cite{liang_2022} for instance.

\section{Flexible Tail Transformation} \label{ftt}

We propose the transformation $R: \mathbb{R} \to \mathbb{R}$ given by
\begin{equation} \label{eq:R2R}
R(z; \mu, \sigma, \lambda_+, \lambda_-) = \mu + \sigma \frac{s}{\lambda_s}[\erfc(|z| / \sqrt{2})^{-\lambda_s} - 1].
\end{equation}
Here $\erfc$ is the \textbf{complementary error function}, a special function reviewed in Appendix \ref{sec:erfc}.
Also $s = \sign(z)$, $\mu$ and $\sigma>0$ control location and scale,
and $\lambda_+ > 0, \lambda_- > 0$ (shorthand for $\lambda_{+1}, \lambda_{-1}$) control tail weights for the upper and lower tails,
allowing us to capture tail weight asymmetry in these two tails.

Most normalising flows use a Gaussian base distribution and a Lipschitz transformation.
Therefore they have Gaussian tails, by the result of \cite{jaini_2020}
described in Section \ref{sec:related_work}.
A property of \eqref{eq:R2R} is that if $X$ has Gaussian tails, then $R(X)$ has GPD tails.
So the transformation can be composed with existing normalising flows to produce heavy tails with parameterised weights.

The transformation \eqref{eq:R2R} is intended to be a final transformation, which adjusts location, scale and tail behaviour.
Preceding transformations can capture further detail in the body of the distribution.
We hypothesise that capturing the extremes in the final transformation will make the downstream flow easier to learn.

To evaluate the likelihood of observations via \eqref{eq:change_of_variables} we need to evaluate the inverse and derivative of \eqref{eq:R2R}, which are provided in Appendix \ref{sec:tf_details} alongside more on the derivation and properties of the proposed transformation.
In the remainder of this section we describe useful modifications to \eqref{eq:R2R}.

\paragraph{Non-GPD Tails}\mbox{}\\
As mentioned, if $X$ has Gaussian tails, then $R(X)$ has GPD tails.
We would like to also permit output with lighter tails, such as Gaussian tails.
Thus we extend $R$ so that negative $\lambda$ parameters produce power transformations,
including an identity transformation.
For full details, see Appendix \ref{sec:nonGPD}.
The functional form is designed so that $R$ remains a smooth function of $z$ regardless of the $\lambda$ values.
However $R$ has a discontinuity as a function of $\lambda$ as it changes between the two regimes.
Theoretically this could cause problems in optimisation, but we have not noticed any empirically.

Even after this modification, our transformation cannot produce tails which are lighter than Gaussian,
for instance tails with bounded support such as uniforms.

\paragraph{Multivariate Transformation}\mbox{}\\
For density estimation, we want to compute the inverse   transformation $R^{-1}(x)$ on $x \in \mathbb{R}^d$ quickly.
For a univariate transformation, let $h = (\mu, \sigma, \lambda_+,\lambda_-)$.
To achieve a multivariate transformation, we perform elementwise transformations $z_i = R^{-1}(x_i; h_i)$.
We use a standard approach to select the $h_i$s:
in the terminology of \cite{papamakarios_2021}, an autoregressive flow with a masked conditioner.
In brief, we take $h_i = c_i(x_{<i})$, so $h_i$ is a function of $x_j$ values with $j<i$.
The transformation $c_i$ is a neural network, using masking so that all $h_i$s can be evaluated from a forward pass of a single network.
This configuration is common practice in density estimation applications.
This approach allows tail behaviour to change in different parts of the distribution, potentially capturing more complex distributional structures than fixed marginal transformations.
On the other hand, classic neural network literature \cite{lecun_2002} argues that neural network input should be standardised to have mean zero and standard deviation one.
As we believe that $x$ has heavy tails, this could negatively impact the ability of the conditioner network to successfully learn the optimal $h$.
As such, we also consider performing the transformation for each marginal independently.

\section{Experiment} \label{sec:experiments}

The central aim of our experimentation was to demonstrate the feasibility of our proposal in modelling high dimensional return data.
We ran experiments targeting a varying number of dimensions $d \in \{10, 100, 200, 300\}$, with data covering the time period 2010-01-04 to 2022-10-27 inclusive.
For a given $d$, we take the closing prices of the top $d$ most traded S\&P 500 stocks, and convert them in standard fashion to log returns.
Specifically, for a single stock closing price $S_j$ with value $S_{j+1}$ the following day, we take $x_j = \log(\frac{S_{j+1}}{S_j})$.
In this work we are concerned with the tails of the data, rather than time series structure.
As such, we treat each day of log returns as an independent observation in $\mathbb{R}^d$.

Preliminary analysis confirmed that the resulting data does exhibit heavy tails in the marginal distributions.
Our expectation is that to achieve good fits we will need to capture extremes in the NF.

\paragraph{Flow Architectures}\mbox{}\\
We compare a number of normalising flow architectures.
The first is gTAF \cite{laszkiewicz_2022}, which uses independent Student's Ts as its base distribution.
For the gTAF transformation we first apply an autoregressive rational quadratic spline (RQS) layer \cite{durkan_2019}, followed by an autoregressive affine layer.
The second is a flow which doesn't explicitly deal with extremes, referred to as RQS.
This is the same as gTAF except it has a $\mathcal{N}(0,I)$ base distribution.

We compare these baselines to 3 variations of our approach, each using a $d$ dimensional standard Gaussian as the base distribution, to which we apply an autoregressive RQS layer, followed by some form of our  tail transformation outlined in Section \ref{ftt}.
The 3 versions we consider are:
(1) TTF (tail transforming flow), which allows non-GPD tails and a multivariate transformation, as described at the end of Section \ref{ftt};
(2) TTF\_m, which differs from TTF by performing a less general multivariate transformation, simply applying a marginal transformation to each component of the input;
(3) EXF (extreme flow), which differs from TTF by not allowing non-GPD tails (i.e.~no negative values of $\lambda$).
All of the considered architectures include trainable linear layers, which form a super-set of permutations and are optimised alongside the other transformation parameters as standard.
Following the definition of gTAF, as described in \cite{laszkiewicz_2022}, we use an LU parametrization \cite{oliva_2018a}.
In the case of gTAF this layer is between the base distribution and the rest of the flow, in other methods the layer is precedes the final layer.
The central difference between gTAF and variants of our approach is where they capture the extremes, in the base distribution or in the final layer.
More details of the experiments are provided in Appendix \ref{sec:implementation_details}.

\paragraph{Experimental Details}\mbox{}\\
We run 10 repeats for each flow/target combination.
Each trains for 300 epochs of the Adam optimiser using a learning rate of 1e-3.
We perform a 40/20/40 split on the data, to give training, validation and test sets respectively.
The test set is comprised of observations after 2017-09-14, with train and validation sampled uniformly from the period up to and including this date.
The date was chosen arbitrarily to give the above proportions of data.
We perform a standard early stopping procedure, with the final model chosen as that which achieved the lowest loss on the validation set.

\begin{figure}
\includegraphics[width=\textwidth]{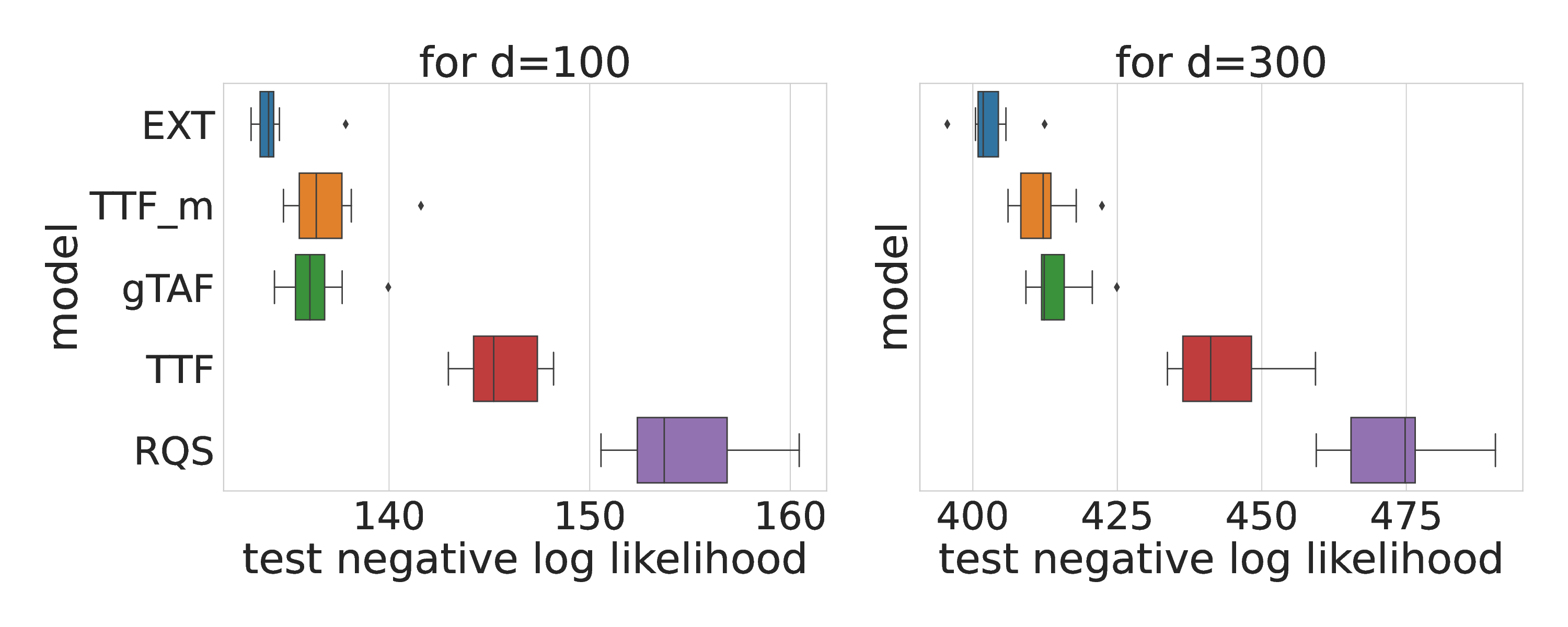}
\caption{Box plots of observed test loglikehoods over 10 repeats.} \label{fig1}
\end{figure}

\paragraph{Goodness of Fit}\mbox{}\\
Table \ref{tab:funnel} reports the negative log likelihoods of the trained models on the test set (visually in figure \ref{fig1}).
The results demonstrate the need to capture extremes in the data, with the RQS model performing consistently poorly.
Our transformation only for GPD tails (EXF) performs the best.
Although TTF is a superset of EXF, so in theory should perform at least as well, it appears that the optimisation is more difficult.
The version of TTF which performs marginal transformations also performs significantly better, somewhat supporting the approach of modelling marginals in a separate procedure as suggested in \cite{mcDonald_2022}.
Although EXF performs best, application of gTAF \cite{laszkiewicz_2022} does provide significant improvements over a naive application of NFs to this data.

\captionof{table}{
Results of density estimation.
Each entry is a mean value across 10 repeated experiments, alongside the standard error in brackets.
The best result for each $d$ is highlighted in bold.
} \label{tab:funnel}
\begin{center}
\begin{tabular}{ccc}
    $d$ & Flow & Negative LL \\ \hline
    \multirow[c]{5}{*}{10}
        & EXF & \textbf{14.38 (0.02)} \\
        & TTF\_m & 14.63 (0.05) \\
        & gTAF & 14.50 (0.05) \\
        & TTF & 14.73 (0.05) \\
        & RQS & 15.16 (0.07) \\

    \hline
    \multirow[c]{5}{*}{100}
        & EXF & \textbf{134.25 (0.42)} \\
        & TTF\_m & 139.27 (0.57) \\
        & gTAF & 136.28 (0.51) \\
        & TTF & 145.53 (0.63) \\
        & RQS & 154.66 (1.05) \\
 \hline
\end{tabular}
\qquad\qquad 
\begin{tabular}{ccc}
    $d$ & Flow & Negative LL \\
    \hline
    \multirow[c]{5}{*}{200}
        & EXF & \textbf{267.99 (1.53)} \\
        & TTF\_m & 279.40 (1.44) \\
        & gTAF & 276.18 (1.06) \\
        & TTF & 292.58 (1.52) \\
        & RQS & 316.17 (1.91) \\

    \hline
    \multirow[c]{5}{*}{300}
        & EXF & \textbf{402.94 (1.54)} \\
        & TTF\_m & 419.15 (2.14) \\
        & gTAF & 414.40 (1.59) \\
        & TTF & 442.94 (2.62) \\
        & RQS & 473.28 (3.08) \\
 \hline
\end{tabular}
\end{center}

\section{Conclusion}

Current methods for modelling extremes with normalising flows use heavy tailed base distributions.
We present an alternative using a flexible final transformation which can induce a tunable range of heavy tail weights.
We explore several variants of this approach when fitting financial returns data, and show one has a clear advantage.

Further work is required to understand why this advantage arises and provide concrete recommendations for a best approach.
One explanation is that by capturing extremeness in the final transformation, the downstream flow is easier to learn.
The advantage of the simpler marginal transformation layer TTF\_m over the more complex auto regressive TTF supports our concerns around passing extreme values to neural networks.
On the other hand, the superior performance of EXF which does exactly that suggests this isn't an issue in all cases.

Anther clear avenue for future work is to incorporate temporal information into the flow models, by incorporating lagged features as inputs to the conditioner, as has been done in other work \cite{wiese2021multiasset}.

\paragraph{Acknowledgements}\mbox{}\\
Thanks to Iain Murray and Jenny Wadsworth for helpful discussions.
Tennessee Hickling is supported by a PhD studentship from the EPSRC Centre for Doctoral Training in Computational Statistics and Data Science (COMPASS). 

\newpage
\appendix
\section{Complementary error function} \label{sec:erfc}

For $z \in \mathbb{R}$, the error function and complementary error function are defined as
\begin{align*}
\erf(z) &= \frac{2}{\sqrt \pi} \int_0^z \exp(-t^2) dt, \\
\erfc(z) &= 1 - \erf(z).
\end{align*}
For large $z$, $\erfc(z) \approx 0$, so it can be represented well in floating point arithmetic.

Below we will use the result that
\begin{equation} \label{eq:f_and_erf}
F(z) = \frac{1}{2}(1 + \erf[z/\sqrt{2}]).
\end{equation}
where $F(z)$ is the $\mathcal{N}(0,1)$ cumulative distribution function.

Efficient numerical evaluation of $\erfc$ and its gradient is possible, as it is a standard special function \cite{temme_2010},
implemented in many computer packages.
For instance PyTorch provides the \texttt{torch.special.erfc} function.
Later we also require $\erfc^{-1}$,
which is less commonly implemented directly.
In this case we apply the result
\[
\erfc^{-1}(x) = -F^{-1}(x/2) / \sqrt{2},
\]
since $F^{-1}$ is a common function.
For instance PyTorch implements it as \texttt{torch.special.ndtri}.

\section{Derivation and properties of transformation} \label{sec:tf_details}

\subsection{Derivation}

Our main transformation \eqref{eq:R2R} can be derived from some simpler transformations motivated by EVT.

\paragraph{GPD transform}

Section \ref{sec:EVT} motivates the transformation $P: [0, 1] \to \mathbb{R}^+$
given by the GPD quantile function
\begin{equation} \label{eq:GPDcdf}
P(u; \lambda) = \frac{1}{\lambda}[(1 - u)^{-\lambda} - 1],
\end{equation}
where $\lambda > 0$ is the tail index.
Using $P$ transforms a distribution with support $[0,1]$ to one on $\mathbb{R}^+$ with tunable tail weight.

We remark that \eqref{eq:GPDcdf} is similar to the popular Box-Cox transformation \cite{box_1964}, although with a restricted domain.
We also note that \eqref{eq:GPDcdf} is also valid for $\lambda < 0$ and maps $[0, 1]$ to $[0, \frac{1}{\lambda}]$.
However this is less useful as a normalising flow transformation as the image depends on the parameter value.

\paragraph{Two tailed transform}

We can extend \eqref{eq:GPDcdf} to a transformation $Q: [-1,1] \to \mathbb{R}$,
\begin{equation} \label{eq:GPD2tails}
Q(u; \lambda_+, \lambda_-) = \frac{s}{\lambda_s}[(1 - |u|)^{-\lambda_s} - 1].
\end{equation}
where $s = \sign(u)$.
Here $\lambda_+ > 0, \lambda_- > 0$ (shorthand for $\lambda_{+1}, \lambda_{-1}$) are tail indices for the positive and negative tails.

Using $Q$ transforms a distribution with support $[-1,1]$ to one on $\mathbb{R}$ with tunable weights for both tails.
In \cite{mcDonald_2022}, a related transformation is applied to margins of a distribution specified by a copula.

\paragraph{Real domain transform}

As discussed in the main paper,
we would like a transformation $R: \mathbb{R} \to \mathbb{R}$ which can transform Gaussian tails to GPD tails.
Consider $z \in \mathbb{R}$,
and let $u = 2F(z)-1$, where $F$ is the $\mathcal{N}(0,1)$ cumulative distribution function.
Then $u \in [-1,1]$ and we can output $Q(u)$.
A drawback is that large $|z|$ can give $u$ values which are rounded to $\pm1$ numerically.

Using \eqref{eq:f_and_erf} shows that
\begin{equation} \label{eq:erfc_relation}
1-|u| = \erfc(|z|/\sqrt{2}),
\end{equation}
so there is a standard special function to compute $1-|u|$ directly.
Large $|z|$ gives $1-|u| \approx 0$ which is less susceptible to rounding issues.

Substituting \eqref{eq:erfc_relation} into \eqref{eq:GPD2tails}, and adding location and scale parameters,
results in our proposed transformation \eqref{eq:R2R}.

\subsection{Properties}

\paragraph{Forward transformation}
Recall our forward transformation is
\begin{equation}
R(z; \mu, \sigma, \lambda_+, \lambda_-) = \mu + \sigma \frac{s}{\lambda_s}[\erfc(|z| / \sqrt{2})^{-\lambda_s} - 1],
\end{equation}
where $s = \sign(z)$.

The derivative of $R$ with respect to $z$ is given by
\begin{equation}
    \frac{\partial R}{\partial z}(z; \mu, \sigma, \lambda_+, \lambda_-) = \sigma \sqrt{\frac{2}{\pi}}\exp(-z^2 / 2) \erfc(|z| / \sqrt{2})^{-\lambda_s -1}.
\end{equation}
All of these terms are positive, which implies that for all parameter settings  $\frac{\partial R}{\partial z}(z; \mu, \sigma, \lambda_+, \lambda_-) > 0$ and the transformation is monotonically increasing.

We can confirm that the transformation is smooth at $z=0$ as
\begin{equation}
    \frac{\partial R}{\partial z}(0; \mu, \sigma, \lambda_+, \lambda_-) = \sigma \sqrt{\frac{2}{\pi}},
\end{equation}
which has no dependence on the tail parameters and is the limit as $z \to 0$ from above or below.

\paragraph{Inverse transformation}
Define $y := \lambda_s |(x-\mu)/\sigma| + 1$.
Then the inverse transformation is
\begin{equation}
R^{-1}(x; \mu, \sigma, \lambda_+, \lambda_-) = s\sqrt{2}\erfc^{-1}(y^{-1/\lambda_s}),
\end{equation}
and its gradient is
\begin{equation}
\frac{\partial R^{-1}}{\partial x}(x; \mu, \sigma, \lambda_+, \lambda_-) = \frac{1}{\sigma}\sqrt{\frac{\pi}{2}} y^{-1/\lambda_s -1} \exp \left( \erfc^{-1}( y^{-1/\lambda_s})^2 \right).
\end{equation}

\subsection{Non-GPD tails} \label{sec:nonGPD}

For $\xi>0$ and $z \geq 0$, define the power transformation
\[
S(z; \xi) = \sqrt{\frac{2}{\pi}}[(1 + z/\xi)^\xi - 1].
\]
For $\xi = 1$ this is simply the identity, with increasingly heavy tails as $\xi$ increases.
Also note that $S(0; \xi) = 0$ and $\frac{dS}{dz} \big|_{z=0} = \sqrt{\frac{2}{\pi}}$,
giving continuity at $z=0$ between the tail and power transformations and between their derivatives.

We extend our main transformation as follows:
\[
\tilde{R}(z; \mu, \sigma, \lambda_+, \lambda_-) =
\begin{cases}
\mu + \sigma \frac{s}{\lambda_s}[\erfc(|z| / \sqrt{2})^{-\lambda_s} - 1]
& \text{for } \lambda_s > 0 \\
\mu + \sigma s S(|z|; \lambda_s + 2)
& \text{for } -1 \leq \lambda_s < 0 \\
\end{cases}
\]

\section{Implementation Details} \label{sec:implementation_details}
We use the \texttt{nflows} package \cite{nflows} to implement TTF.
Both of these libraries depend on pytorch \cite{pytorch} for automatic differentiation.

In both gTAF and TTF, our autoregressive layers are configured to produce the required transformation parameters for each input via the forward pass of masked neural networks.
We use a default of 2 hidden layers, each with a width calculated as the dimension of the problem + 10.

The RQS layers are configured with a bounding box of $[-2.5, 2.5]$ and $8$ knot points.

Full details can be found in our code for these examples at 
\url{https://github.com/tennessee-wallaceh/tailnflows}.

\newpage
\bibliographystyle{splncs04}  
\bibliography{references} 

\begin{thebibliography}{10}
\providecommand{\url}[1]{\texttt{#1}}
\providecommand{\urlprefix}{URL }
\providecommand{\doi}[1]{https://doi.org/#1}

\bibitem{amiri_2022}
Amiri, S., Nalisnick, E.T., Belloum, A., Klous, S., Gommans, L.: Generating heavy-tailed synthetic data with normalizing flows (2022), \url{https://openreview.net/forum?id=PbvyJ8XpNn}

\bibitem{box_1964}
Box, G.E.P., Cox, D.R.: An analysis of transformations. Journal of the Royal Statistical Society. Series B (Methodological)  (1964)

\bibitem{buehler_2018}
Bühler, H., Gonon, L., Teichmann, J., Wood, B.: Deep hedging (2018), arXiv: 1802.03042

\bibitem{cirillo_2020_disease_risk}
Cirillo, P., Taleb, N.N.: Tail risk of contagious diseases. Nature Physics  \textbf{16},  606--613 (2020)

\bibitem{coles_2001}
Coles, S.: An Introduction to Statistical Modeling of Extreme Values. Springer London, London (2001)

\bibitem{durkan_2019}
Durkan, C., Bekasov, A., Murray, I., Papamakarios, G.: Neural spline flows. In: Advances in Neural Information Processing Systems. vol.~32 (2019)

\bibitem{nflows}
Durkan, C., Bekasov, A., Murray, I., Papamakarios, G.: {nflows}: normalizing flows in {PyTorch} (Nov 2020)

\bibitem{foss_2011}
Foss, S., Korshunov, D., Zachary, S.: An introduction to heavy-tailed and subexponential distributions. Springer (2011)

\bibitem{gilli_2006_financial_risk}
Gilli, M., K{\"e}llezi, E.: An application of extreme value theory for measuring financial risk. Computational Economics  \textbf{27},  207--228 (2006)

\bibitem{jaini_2020}
Jaini, P., Kobyzev, I., Yu, Y., Brubaker, M.: Tails of {L}ipschitz triangular flows. In: Proceedings of the 37th International Conference on Machine Learning. vol.~119, pp. 4673--4681. PMLR (2020)

\bibitem{kingma_2018}
Kingma, D.P., Dhariwal, P.: Glow: Generative flow with invertible 1x1 convolutions. In: Advances in Neural Information Processing Systems. vol.~31 (2018)

\bibitem{kobyzev_2020}
Kobyzev, I., Prince, S.J.D., Brubaker, M.A.: Normalizing flows: An introduction and review of current methods. IEEE transactions on pattern analysis and machine intelligence  \textbf{43}(11),  3964--3979 (2020)

\bibitem{laszkiewicz_2022}
Laszkiewicz, M., Lederer, J., Fischer, A.: Marginal tail-adaptive normalizing flows. In: International Conference on Machine Learning. vol.~162, pp. 12020--12048. {PMLR} (2022)

\bibitem{lecun_2002}
LeCun, Y., Bottou, L., Orr, G.B., M{\"u}ller, K.R.: Efficient backprop. In: Neural networks: Tricks of the trade, pp. 9--50. Springer (2002)

\bibitem{liang_2022}
Liang, F.T., Hodgkinson, L., Mahoney, M.W.: Fat-tailed variational inference with anisotropic tail adaptive flows. In: International Conference on Machine Learning. vol.~162, pp. 13257--13270. {PMLR} (2022)

\bibitem{mcDonald_2022}
McDonald, A., Tan, P.N., Luo, L.: Comet flows: Towards generative modeling of multivariate extremes and tail dependence. In: International Joint Conference on Artificial Intelligence (2022)

\bibitem{oliva_2018a}
Oliva, J., Dubey, A., Zaheer, M., Poczos, B., Salakhutdinov, R., Xing, E., Schneider, J.: Transformation autoregressive networks. In: International Conference on Machine Learning. vol.~80, pp. 3898--3907. PMLR (2018)

\bibitem{papamakarios_2021}
Papamakarios, G., Nalisnick, E., Rezende, D.J., Mohamed, S., Lakshminarayanan, B.: Normalizing flows for probabilistic modeling and inference. Journal of Machine Learning Research  \textbf{22}(57),  1--64 (2021)

\bibitem{papastathopoulos_2013}
Papastathopoulos, I., Tawn, J.A.: Extended generalised {P}areto models for tail estimation. Journal of Statistical Planning and Inference  \textbf{143}(1),  131--143 (2013)

\bibitem{pytorch}
Paszke, A., Gross, S., Massa, F., Lerer, A., Bradbury, J., Chanan, G., Killeen, T., Lin, Z., Gimelshein, N., Antiga, L., Desmaison, A., Kopf, A., Yang, E., DeVito, Z., Raison, M., Tejani, A., Chilamkurthy, S., Steiner, B., Fang, L., Bai, J., Chintala, S.: Pytorch: An imperative style, high-performance deep learning library. In: Advances in Neural Information Processing Systems 32, pp. 8024--8035. Curran Associates, Inc. (2019)

\bibitem{pickands_1975}
Pickands~III, J.: Statistical inference using extreme order statistics. the Annals of Statistics pp. 119--131 (1975)

\bibitem{prenger_2018}
Prenger, R., Valle, R., Catanzaro, B.: Waveglow: A flow-based generative network for speech synthesis (2018), arXiv:1811.00002

\bibitem{temme_2010}
Temme, N.M.: Error functions, {D}awson's and {F}resnel integrals. In: {NIST} handbook of mathematical functions. Cambridge university press (2010)

\bibitem{wiese_2020}
Wiese, M., Knobloch, R., Korn, R., Kretschmer, P.: Quant {GANs}: deep generation of financial time series. Quantitative Finance  \textbf{20}(9),  1419--1440 (apr 2020)

\bibitem{wiese2021multiasset}
Wiese, M., Wood, B., Pachoud, A., Korn, R., Buehler, H., Murray, P., Bai, L.: Multi-asset spot and option market simulation (2021), arXiv:2112.06823

\bibitem{zscheischler_2018_climate_risk}
Zscheischler, J., Westra, S., van~den Hurk, B., Seneviratne, S.I., Ward, P.J., Pitman, A.J., Aghakouchak, A., Bresch, D.N., Leonard, M., Wahl, T., Zhang, X.: Future climate risk from compound events. Nature Climate Change  \textbf{8},  469--477 (2018)

\end{thebibliography}

\end{document}